\documentclass[10pt,twocolumn,letterpaper]{article}

\usepackage[pagenumbers]{cvpr} 

\usepackage{graphicx}
\usepackage{amsmath}
\usepackage{amssymb}
\usepackage{booktabs}
\usepackage{bm}
\usepackage{amssymb}
\usepackage{pifont}
\usepackage[pagebackref,breaklinks,colorlinks]{hyperref}
\usepackage{xcolor,colortbl}
\newcommand{\gray}[1]{\textcolor{gray}{#1}}
\definecolor{ffe1da}{RGB}{255,225,218}
\definecolor{F7E0D5}{RGB}{247,224,213}
\definecolor{myred}{RGB}{199,100,38}
\definecolor{darkF7E0D5}{RGB}{209,154,128}
\definecolor{White}{RGB}{255,255,255}
\colorlet{Light}{White!80!myred}
\def\OURS{LOCA\xspace}

\usepackage[capitalize]{cleveref}
\crefname{section}{Sec.}{Secs.}
\Crefname{section}{Section}{Sections}
\Crefname{table}{Table}{Tables}
\crefname{table}{Tab.}{Tabs.}

\newcommand\blfootnote[1]{%
  \begingroup
  \renewcommand\thefootnote{}\footnote{#1}%
  \addtocounter{footnote}{-1}%
  \endgroup
}

\def\cvprPaperID{3660} 
\def\confName{CVPR}
\def\confYear{2023}

\begin{document}

\title{Location-Aware Self-Supervised Transformers for Semantic Segmentation}

\author{Mathilde Caron
~~~~Neil Houlsby
~~~~Cordelia Schmid
\\
Google Research
}
\maketitle

\begin{abstract}
\blfootnote{
Correspondence: \href{mailto:mcaron@google.com}{mcaron@google.com}. Code released at: \\
\href{https://github.com/google-research/scenic/tree/main/scenic/projects/}{https://github.com/google-research/scenic/tree/main/scenic/projects/loca}
}
Pixel-level labels are particularly expensive to acquire.
Hence, pretraining is a critical step to improve models on a task like semantic segmentation.
However, prominent algorithms for pretraining neural networks use image-level objectives, e.g. image classification, image-text alignment à la CLIP, or self-supervised contrastive learning.
These objectives do not model spatial information, which might be sub-optimal when finetuning on downstream tasks with spatial reasoning.
In this work, we pretrain network with a \textbf{loc}ation-\textbf{a}ware (\OURS) self-supervised method which fosters the emergence of strong dense features.
Specifically, we use both a patch-level clustering scheme to mine dense pseudo-labels and a relative location prediction task to encourage learning about object parts and their spatial arrangements.
Our experiments show that \OURS pretraining leads to representations that transfer competitively to challenging and diverse semantic segmentation datasets.
\end{abstract}


\section{Introduction}
\label{sec:intro}

The spatial annotations required for training semantic segmentation models are extremely time consuming and costly to acquire.
Therefore, pretraining is commonly used to improve performance and label-efficiency of these models~\cite{strudel2021segmenter}.
The dominant method for pretraining neural networks uses image-level tasks on massive amounts of supervised data~\cite{russakovsky2015imagenet,radford2021learning,carreira2017quo,zhai2022scaling,yuan2021florence}.
For example, powerful foundation models such as Flamingo~\cite{alayrac2022flamingo}, CoCa~\cite{yu2022coca} or PaLI~\cite{chen2022pali}, build upon a visual encoder pretrained by matching aligned image and text pairs with a contrastive loss~\cite{radford2021learning}, or by classifying images into a predefined set of categories~\cite{zhai2022scaling}.
These two standard supervised pretraining objectives operate at the global (whole image) level, without explicitly encouraging spatial reasoning.

\begin{figure}[t]
\centering
\includegraphics[width=1\linewidth]{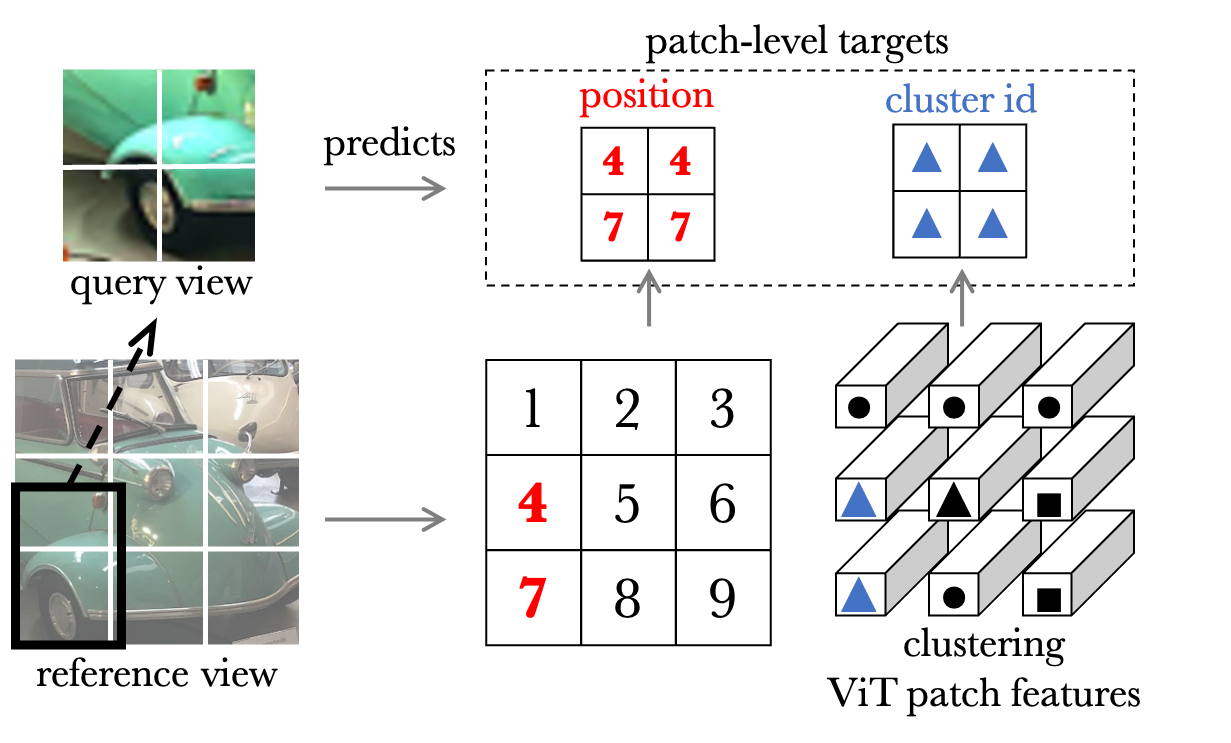}
\vspace{-0.7cm}
\caption{\textbf{\OURS} is a self-supervised pretraining method which combines relative position and patch-level cluster prediction.
This achieves improved transfer on semantic segmentation datasets.
The method is presented in Sec.~\ref{sec:method}.
\vspace{-0.6cm}
}
\label{fig:pullfig}
\end{figure}

However, it is unclear whether image-level pretraining is the optimal strategy when targeting recognition tasks with spatial understanding such as semantic segmentation.
In fact, a recent study by Minderer~\etal~\cite{minderer2022simple} shows that some models pretrained with image classification, while being excellent at image-level downstream tasks, transfer poorly to object detection, a task also requiring spatial reasoning.
We argue that the main reason why pretraining is usually done with global objectives is because annotations are much easier to collect at the image level rather than at the pixel level.
Indeed, the image classification or image-text datasets typically used in state-of-the-art systems~\cite{zhai2022scaling,radford2021learning} are orders of magnitude bigger and cover more categories than densely annotated datasets~\cite{zhou2017scene,lin2014microsoft}.
Therefore, one approach to unlock the potential of dense, spatially-aware pretraining at scale might be to move away from annotations, as proposed by self-supervised learning (SSL) approaches.
A successful branch of SSL, often coined as ``contrastive learning'', works by matching the representation of different views obtained from a same image by means of data augmentation~\cite{he2020momentum,chen2020simple,caron2020unsupervised,grill2020bootstrap}.
Interestingly, Caron~\etal~\cite{caron2021emerging} have shown that segmentation masks emerge from the attention maps of Vision Transformers (ViT)~\cite{dosovitskiy2020image} trained with these contrastive methods and several works have built on this observation to generate completely unsupervised segmentations~\cite{hamilton2022unsupervised,ziegler2022self,simeoni2021localizing}.
However, we found in our preliminary experiments that salient attention maps do not correlate with superior performance after \textit{finetuning} to the semantic segmentation task~\cite{ziegler2022self}.
We hypothesize that this is because contrastive methods operate at the global level without explicitly encouraging spatial relationships.
Worse, due to their intensive use of spatial data augmentation such as cropping and rescaling, they tend to produce \textit{localization-invariant} features discarding spatial information~\cite{chen2022intra,assran2022masked}.

In order to foster the emergence of strong dense representations, our goal is to design a patch-level pretext task encouraging spatial localization reasoning.
Recently, patch-level SSL pretrainings have attracted more and more attention in the community~\cite{he2022masked,bao2021beit,zhai2022position,baevski2022data2vec,xie2022simmim,weinzaepfel2022croco}.
For example, dense contrastive approaches adapt the popular contrastive SSL paradigm to the patch level~\cite{pinheiro2020unsupervised,wang2021dense,roh2021spatially,xiao2021region,xie2021propagate} while masked autoencoders propose to reconstruct masked patches~\cite{he2022masked,bao2021beit}.
Of particular interest, Zhai~\etal~\cite{zhai2022position} propose a pure localization method, that of predicting the patches position in an image.
Intuitively, position prediction should inherently require a strong spatial and semantic understanding and has been the core motivation of the pioneering SSL branch of ``jigsaw puzzle''~\cite{doersch2015unsupervised,noroozi2016unsupervised}.
In this work, we propose to revisit this strategy and introduce a relative position prediction task.
Specifically, our method works by predicting the location of a \textit{query} view relatively to another, \textit{reference}, view.
To be able to locate themselves in the reference, the query patch features ``look'' at those of the reference through shallow cross-attention.
We control the difficulty of the task and properties of the resulting features by masking reference patch features visible to the query.
Our experiments show that this query-reference mechanism improves greatly over the single-view design of Zhai~\etal~\cite{zhai2022position} when transferring to semantic segmentation.

Since semantic segmentation is a per-patch classification problem, we also propose to prepare the ViT features for this task by means of clustering-based pseudo-labeling~\cite{caron2018deep,asano2019self,yang2016joint} done at the patch level~\cite{ziegler2022self}. Overall, we present in this work a \textbf{loc}ation-\textbf{a}ware (\OURS) self-supervised pretraining approach for semantic segmentation, which combines a straightforward patch-level SSL clustering method and relative position pretraining, as illustrated in Fig.~\ref{fig:pullfig}.
We show that \OURS yields improved performance over state-of-the-art supervised~\cite{touvron2022deit,steiner2021train,radford2021learning} and unsupervised~\cite{caron2021emerging,he2022masked,zhou2021ibot,chen2021empirical} representation learning methods for ViTs when transferred to 11 diverse and challenging semantic segmentation benchmarks.
Our method scales promisingly to large models and large amount of data, which is a positive signal that it could be a viable candidate for spatially-aware pretraining at scale.
Finally, we present a thorough analysis of different design choices which have led to the development of \OURS.

\section{Related Work}
\label{sec:related}

\vspace{-0.1cm}
\paragraph{Clustering-based SSL} consists in using clustering while training to mine pseudo-labels in a dataset without annotations~\cite{caron2018deep,yang2016joint}.
This pseudo-assignment strategy is usually done at the image level~\cite{caron2018deep,yang2016joint,caron2020unsupervised,assran2022masked,asano2019self} but recent works propose to cluster patch-level representations~\cite{ziegler2022self,cho2021picie}.
In particular, our work takes inspiration from Leopart~\cite{ziegler2022self} which propose a patch-level cluster prediction task.
However, unlike Leopart, our work leverages an explicit position-based pretraining and simplifies the clustering pipeline.
For example, we omit foreground attention-based pooling, a design choice which makes Leopart a method that starts from an already good backbone~\cite{caron2021emerging} while ours trains from scratch.

\vspace{-0.4cm}
\paragraph{SSL with location prediction.}
Pioneering works in SSL proposed to exploit spatial cues to generate pretext tasks~\cite{doersch2015unsupervised,kim2018learning,lee2017unsupervised,mundhenk2018improvements,santa2017deeppermnet,noroozi2016unsupervised,goyal2019scaling}.
Notably, inspired by word2vec~\cite{mikolov2013efficient}, Doersch~\etal~\cite{doersch2015unsupervised} train a network to predict the relative position of a pair of patches from the same image while Noroozi and Favaro~\cite{noroozi2016unsupervised} extend this approach to solving ``jigsaw puzzles'' by rearranging a set of shuffled crops of an image.
These approaches were developed with Convnets and very little work has revisited them in the scope of Transformers~\cite{zhai2022position}.
Zhai~\etal~\cite{zhai2022position} propose to pretrain a ViT to predict the position of its input patches given their visual appearance only, i.e. by discarding positional embeddings.
We compare this strategy to the \OURS mechanism in Fig.~\ref{fig:query} and Sec.~\ref{sec:localization}.
Also using localization, UP-DETR~\cite{dai2021up} propose to pretrain the DETR~\cite{carion2020end} architecture without using any annotations by localizing random boxes in a reference image.
Our initial design is inspired by UP-DETR, though our implementations vary greatly: we formulate our task as a patch position classification problem while they follow DETR losses and architecture.

\vspace{-0.4cm}
\paragraph{Context and masked auto-encoders.}
Also exploiting spatial cues for SSL, Pathak~\etal~\cite{pathak2016context} propose context auto-encoders to train Convnets to generate the content of a masked region based on its surrounding.
Recently, masked auto-encoders revisit this ``inpainting'' approach to pretraining Vision Transformers~\cite{bao2021beit,he2022masked,wei2022masked}.
Specifically, the task is to reconstruct masked~\cite{bao2021beit} or dropped~\cite{he2022masked} patches from the input sequence tokens, either directly in pixel space~\cite{he2022masked} or in feature space~\cite{wei2022masked,bao2021beit,zhou2021ibot}.
Similar to our approach, masked auto-encoders are trained with patch-based objectives with a task encouraging the network to learn local representations of object parts and their spacial arrangement.

\vspace{-0.4cm}
\paragraph{Dense contrastive SSL.}
A prominent line of SSL, often referred to as ``contrastive'' or ``siamese'' approaches, trains networks by matching the representation of different views obtained from a same image by means of data augmentation~\cite{he2020momentum,chen2020simple,caron2020unsupervised,grill2020bootstrap,wu2018unsupervised,dosovitskiy2016discriminative,assran2022masked}.
These approaches have primarily been developed with global (image-level) objectives but several recent works have adapted them to learn local features~\cite{pinheiro2020unsupervised,wang2021dense,roh2021spatially,xiao2021region,xie2021propagate,yun2022patch,huang2022learning}.
Specifically, instead of matching representations from global descriptors, they match features that come from the same location in the original image but seen from different views~\cite{pinheiro2020unsupervised}.
We borrow the strategy of back-tracking the data augmentation process of two views to find their regions that intersect~\cite{pinheiro2020unsupervised}.

\vspace{-0.4cm}
\paragraph{Unsupervised semantic segmentation.}
While our goal is to improve pretraining for semantic segmentation, some parallel works to ours directly target semantic segmentation without using any supervision at all~\cite{cho2021picie,van2021unsupervised}.
Indeed, Caron~\etal~\cite{caron2021emerging} have shown that unsupervised segmentation masks emerge from the attention module of ViTs trained with image-level contrastive objectives such as DINO.
Several works build on this observation and enhance SSL features to produce completely unsupervised segmentations~\cite{hamilton2022unsupervised,ziegler2022self,simeoni2021localizing}.
These approaches typically do not evaluate semantic segmentation after end-to-end finetuning.

\section{Methodology}
\label{sec:method}
\vspace{-0.1cm}
In order to foster the emergence of strong dense representations for semantic segmentation, we design \OURS, a patch-level SSL pretraining task that requires to reason about spatial localization.
It works as a query-reference scheme where patches of a query view predict both their position and their cluster assignment relatively to a reference view, as illustrated in Fig~\ref{fig:pullfig}.

\vspace{-0.4cm}
\paragraph{Generating query and reference views.}
From an image $x$ of a dataset, we form a \textit{reference} view (denoted by $\bm{x}_{ref}$) and a \textit{query} view (denoted by $\bm{x}_q$) using a randomized data augmentation routine composed of flipping, cropping, rescaling and color jittering.
Because query and reference are generated by two independent augmentation draws, they usually have different image statistics (i.e. different scale, region or color histogram).
This forces the network to rely less on low-level cues (chromatic aberration, color, and edge consistency) to solve the self-supervised task and more on recognizing object parts and their organization.

The query's predictions are supervised by the reference view and therefore our loss is defined only at the intersection of the two views.
Hence, we want the query and reference to intersect often.
Also, we wish to constrain the spatial extent of the queries in order to favor the emergence of image-\textit{part} representations.
A natural choice then is to sample the reference view so that it covers a large area of the original image and the query views so that they cover a small portion of the original image.
In practice we use the MSN input pipeline~\cite{assran2022masked} with random resize-cropping and patch-dropping for generating different query views per reference.
We consider a single query when describing the method for simplicity but use ten in our experiments.

\vspace{-0.4cm}
\paragraph{Correspondences between query and reference.}
Following the standard protocol of the Vision Transformers~\cite{dosovitskiy2020image}, query and reference views are divided into non overlapping patches of resolution $P \times P$.
More precisely, the reference view is flattened into $N_{ref} = \lfloor H_{ref} / P\rfloor  \times \lfloor W_{ref} / P\rfloor $ (with $H_{ref} \times W_{ref}$ the resolution of $x_{ref}$) separate patches $\bm{x}_{ref}^i, i \in \{1, ..., N_{ref}\}$.
Default typical values are $H_{ref} = W_{ref} = 224$, $P = 16$ and $N_{ref} = 196$.
A similar ``patchification'' process is applied on the query view, resulting in a sequence of $N_q$ patches $\bm{x}_{q}^j, j \in \{1, ..., N_{q}\}$.
Unless specified otherwise, we have $N_q = 36$.
By back-tracking the data augmentation draws that generated $\bm{x}_{ref}$ and $\bm{x}_q$, we can identify the patch-level correspondences between these two views.
In particular, we know a function $\bm{h}$ that, given any patch position $j$ in the query view, returns the position $i = \bm{h}(j)$ of the patch in the reference, $\bm{x}_{ref}^{\bm{h}(j)}$, that has the greatest overlap with the query patch $\bm{x}_{q}^j$.
We implement the function $\bm{h}$ with successive nearest interpolations and because the patchification grids of $\bm{x}_q$ and $\bm{x}_{ref}$ are usually not exactly aligned, a pair of matching patches, $\bm{x}_q^{j}$ and $\bm{x}_{ref}^{\bm{h}(j)}$, have similar content but do not generally match \textit{perfectly}.
This effect can be seen in the example in Fig.~\ref{fig:pullfig}.

\vspace{-0.4cm}
\paragraph{Patch-level encoding with ViT.}
We process both the reference and query views with a Vision Transformer network~\cite{dosovitskiy2020image}, denoted by $f$, of internal dimension $d$ ($d = 768$ for ViT-B).
We note $\bm{Z}_q \in \mathbb{R}^{d \times N_q}$ (resp. $\bm{Z}_{ref} \in \mathbb{R}^{d \times N_{ref}}$), the output patch-level representation matrix of the query (resp. reference) view.
As commonly done in SSL~\cite{chen2020simple,caron2020unsupervised,grill2020bootstrap}, we project these representations with a 2-layer multilayer perceptron (MLP), resulting in features $\bm{\Tilde{Z}}_q \in \mathbb{R}^{\Tilde{d} \times N_q}$ and $\bm{\Tilde{Z}}_{ref} \in \mathbb{R}^{\Tilde{d} \times N_{ref}}$ with $\Tilde{d} = 256$.

\begin{table*}[t]
    \caption{
      \textbf{Comparison with other SSL pretrainings on 11 semantic segmentation benchmarks.} 
We report mean IoU on the different validation sets.
All methods use ImageNet-1k and ViT-B/16.
We use the linear decoder from Segmenter~\cite{strudel2021segmenter} and run evaluation for other methods from their publicly released checkpoints.
In the last column, we report the relative improvement over starting from random init. averaged across the 11 datasets.
\OURS improves over MAE by $+4.3$ points.
}
\vspace{-0.15cm}
\centering
\small
  \setlength{\tabcolsep}{2.3pt}
       \begin{tabular}{@{} l ccc c ccccc c c c c c c | c@{}}
      \toprule
      & \multicolumn{3}{c}{\textit{Consumer}} && \multicolumn{5}{c}{\textit{Driving}} && \multicolumn{1}{c}{\textit{Indoor}} && \multicolumn{1}{c}{\textit{Aerial}} && \multicolumn{1}{c|}{\textit{Underwater}} & \multicolumn{1}{c}{Avg. rel. } \\
\cmidrule{2-4}\cmidrule{6-10}\cmidrule{12-12}\cmidrule{14-14}\cmidrule{16-16}
	    Pretraining method   & ADE20k & P.Cont & P.VOC && Citys. & BDD & CamVid & IDD & KITTI && SUN && ISPRS && SUIM & $\Delta$ (\%)\\
      \midrule
\gray{Random init.} & \gray{21.1} & \gray{19.6} & \gray{29.1} && \gray{51.4} &  \gray{40.2} & \gray{43.3} & \gray{45.2} & \gray{39.0} && \gray{19.7} && \gray{28.1} && \gray{53.0} & \gray{0} \\
\gray{Supervised - \footnotesize{DeiT-III}~\cite{touvron2022deit}} & \gray{47.3} & \gray{53.9} & \gray{76.1} && \gray{79.7} &  \gray{62.7} & \gray{53.8} & \gray{55.4} & \gray{47.2} && \gray{47.5} && \gray{42.1} && \gray{73.5} & \gray{79.0} \\
DINO~\cite{caron2021emerging} & 44.1 & 50.7 & 74.1 && 78.4 & 60.7 & 51.5 & 54.3 & 46.4 && 44.4 && 41.5 && 71.2 & 71.9\\
MoCo-v3~\cite{chen2021empirical} & 45.4 & 51.6 & 74.5 && 78.6 & 60.4 & 51.1 & 53.7 & 45.7 && 45.6 && 42.1 && 72.6 & 73.6 \\
iBOT~\cite{zhou2021ibot} & 47.0 & 54.6 & 75.0 && \textbf{79.8} & 62.1 & 51.5 & 55.5 & 47.0 && 46.3 && 42.2 && 73.2 & 77.7 \\
MAE~\cite{he2022masked} & 45.5 & 51.7 & 75.0 && 79.7 & 62.1 & \textbf{57.8} & \textbf{55.8} & 48.3 && 45.9 && 44.6 && 72.4 & 77.8 \\
\rowcolor{Light}
\OURS (Ours) & \textbf{47.9} & \textbf{54.9} & \textbf{76.7} && \textbf{79.8} & \textbf{62.8} & 56.1 & 55.6 & \textbf{48.5} && \textbf{47.7} && \textbf{45.6} && \textbf{74.0} & \textbf{82.1} \\
\bottomrule
 \end{tabular}
    \label{tab:compare-ssl}
    \vspace{-0.4cm}
\end{table*}

\vspace{-0.4cm}
\paragraph{Patch-level clustering.}
Training for semantic segmentation in a supervised setting is typically cast as a per-patch classification problem over $K$ predefined categories:
$$
\frac{1}{N_{q}} \sum_{j=1}^{N_{q}}  \ell( (Q^T\bm{\Tilde{Z}}_q)_j, y_j)
$$
where $Q$ is a matrix in $\mathbb{R}^{\Tilde{d} \times K}$ of learnable category prototypes and $\ell$ is the softmax cross-entropy loss.
This problem is supervised by patch-level annotations $y_j$.
However, because we do not have access to any annotations, inspired by previous SSL works~\cite{caron2018deep,caron2020unsupervised}, we resort to clustering for pseudo-supervision.
In particular, to supervise the patch $j$ in the query, we cluster the patch representations of another view of the same image, i.e. the reference view.
we obtain a soft cluster assignment (or pseudo-label) based on the similarity between the prototypes and the patch representation at the corresponding localization in the reference view:
$$
\bm{y}_{ref}^{i} = \text{softmax}( \bm{\Tilde{Z}}_{ref}^i . Q / \tau)
$$
with $i = \bm{h}(j)$ and $\tau$ a temperature parameter controlling the sharpness of the distribution.
We further adjust the cluster assignment distribution with Sinkhorn-Knopp~\cite{cuturi2013sinkhorn} to avoid the collapsing trivial solution~\cite{caron2020unsupervised,asano2019self} and to encourage using equally all the clusters.
Now that we have replaced expensive per-patch label supervision with cluster pseudo-labels we can minimize the following objective:
\begin{eqnarray}\label{eq:feat}
\frac{1}{|\Omega|} \sum_{j \in \Omega} \ell( (Q^T\bm{\Tilde{Z}}_q)_j, \bm{y}_{ref}^{\bm{h}(j)})
\end{eqnarray}
where $\Omega$ is the set of patch position in the query that has an intersection with the reference (i.e. where $\bm{h}$ is defined).
We regularize this loss function with the mean entropy maximization (me-max) protocol~\cite{assran2022masked} to encourage the network to use the full set of pseudo-label prototypes $Q$ (see Tab.~\ref{fig:multi_ablat}a)).

\vspace{-0.4cm}
\paragraph{Patch position prediction.}
To encourage the network to learn about different object parts and their spatial arrangement, we propose to predict relative patch positions.
We implement a query localization problem as a $N_{ref}$-way classification task where each query patch representations has to predict the position of the patch covering the same content in the reference view, as given by $\bm{h}$.
To that end, the patch representations of the query need to be able to ``look'' at those of the reference.
We implement this query-reference interaction with a single cross-attention transformer block, denoted by $g$, whose queries are computed from $\bm{Z}_{q}$ and keys and values are obtained from $\bm{Z}_{ref}$.
We denote the query representations after they have looked at the reference as $\bm{G} = g(\bm{Z}_q, \bm{Z}_{ref}) \in \mathbb{R}^{d \times N_q}$ and by $W \in \mathbb{R}^{d \times N_{ref}}$ the final ``position classification'' layer.
Note that $N_{ref}$ is the total number of positions in the reference.
We train the network to minimize the following position prediction loss:
\begin{eqnarray}\label{eq:position}
\frac{1}{|\Omega|} \sum_{j \in \Omega}  \ell( (W^T\bm{G})_j, \bm{h}(j))
\end{eqnarray}
where $\Omega$ and $\ell$ are defined as in Eq~\ref{eq:feat}.

\vspace{-0.4cm}
\paragraph{Masking reference patch features visible to the query.}
In practice, we find that problem~\ref{eq:position} can be solved near perfectly by the network (see the validation accuracy in Fig.~\ref{fig:patch_pos}).
As empirically shown in Sec.~\ref{par:query}, one strategy to make the task more challenging is to restrict what the query can see from the reference.
We implement this mechanism by randomly dropping (or ``masking'') a ratio $\eta$ of the patch features input to the cross-attention block $g$.
Specifically, we redefine $\bm{G} = g(\bm{Z}_q, m(\bm{Z}_{ref}, \eta))$ where $m$ is a random process that discards $\lfloor \eta N_{ref} \rfloor$ columns of $\bm{Z}_{ref}$.
We use structured dropping (i.e. we keep a consecutive subset of patch tokens) as we find in our experiments that it leads to superior performance than unstructured dropping ($+ 0.8$ mIoU).

\vspace{-0.4cm}
\paragraph{Optimization.}
We train \OURS by minimizing the sum of the objectives in Eq~\ref{eq:feat} and Eq~\ref{eq:position}, averaged both over the different query views and the minibatch.
We learn the parameters of $f$, $g$, $Q$, and $W$ by back-propagating in the branch processing the query views.
The parameters used in the branch processing the reference views are updated via an exponential moving average of the encoder parameters processing the query views~\cite{he2020momentum,grill2020bootstrap,caron2021emerging}.
This asymmetry improves performance and stability for clusters prediction and does not have any effect on the position prediction.

\vspace{-0.4cm}
\paragraph{Implementation and evaluation.}
We train \OURS with learning rate of $0.001$ (cosine schedule), batch size of $1024$ and weight decay of $0.1$ with adamw~\cite{loshchilov2018fixing}.
Models in Sec.~\ref{sec:compare} are trained for 600 epochs and those for analyses (Sec.~\ref{sec:design_choices}) for 100 epochs.
We evaluate by end-to-end finetuning on 11 semantic segmentation benchmarks~\cite{mensink2021factors}: ADE20k~\cite{zhou2017scene}, Pascal Context~(``P.Cont'')~\cite{mottaghi2014role}, Pascal VOC~(``P.VOC'')~\cite{everingham2010pascal}, Cityscapes~(``Citys.'')~\cite{cordts2016cityscapes}, Berkeley Deep Drive (``BDD'')~\cite{yu2020bdd100k}, CamVid~\cite{camvid}, India Driving Dataset (``IDD'')~\cite{varma2019idd}, KITTI~\cite{abu2018augmented}, SUN-RGB-D~(``SUN'')~\cite{song2015sun}, ISPRS~\cite{meidow2014theme} and SUIM~\cite{islam2020semantic}.
We follow and reproduce the linear decoder protocol of~\cite{strudel2021segmenter}.
It uses a minimal amount of adapter layers to prevent the effect of pretraining of being washed out by heavy decoders.
We report results for other methods if available and run evaluation from released checkpoints if not.
We run a hyperparameter search with the same budget for all methods.
We report results in single scale, averaged over 5 runs.
All implementation details are in the Appendix~\ref{ap1}.

\section{Main Results}
\label{sec:compare}

\begin{table}[t]
    \caption{
      \textbf{Fewshot semantic segmentation.} 
We report mean IoU on the validation set of ADE20k for different SSL pretrained models.
All methods use ImageNet-1k and ViT-B/16.
Only a fraction of training images are used for finetuning.
Results are averaged over $5$ different splits.
}
\centering
\vspace{-0.15cm}
\small
  \setlength{\tabcolsep}{4pt}
    \begin{tabular}{@{} l c c c c c c @{}}
      \toprule
	    Method      & $1/32$ & $1/16$ & $1/8$ & $1/4$ & $1/2$ & $1$ \\
      \midrule
      \gray{Supervised - \footnotesize{DeiT-III}~\cite{touvron2022deit}} & \gray{20.9} & \gray{27.1} & \gray{32.7} & \gray{38.3} & \gray{42.0} & \gray{47.3} \\
      DINO~\cite{caron2021emerging} & 18.4 & 24.5 & 29.5 & 35.2 & 39.5 & 44.1 \\
      MoCo-v3~\cite{chen2021empirical} & 17.7 & 25.2 & 30.8 & 36.5 & 40.7 & 45.4\\
      iBOT~\cite{zhou2021ibot} & 20.9 & 28.0 & 33.4 & 38.7 & 42.6 & 47.0\\
      MAE~\cite{he2022masked} & 18.4 & 25.3 & 30.5 & 36.1 & 40.6 & 45.5 \\
      \rowcolor{Light}
      \OURS (Ours) & \textbf{22.2} & \textbf{30.0} & \textbf{34.4} & \textbf{39.1} & \textbf{42.8} & \textbf{47.9} \\
      \bottomrule
 \end{tabular}
    \label{tab:fewshot}
    \vspace{-0.4cm}
\end{table}

\vspace{-0.1cm}
\subsection{Comparison with other SSL pretrainings}
\vspace{-0.1cm}
In this section, we compare \OURS to popular state-of-the-art SSL models for ViTs: DINO~\cite{caron2021emerging}, MoCo-v3~\cite{chen2021empirical}, MAE~\cite{he2022masked} and iBOT~\cite{zhou2021ibot}.
The compared models all use ImageNet-1k (without labels) and ViT-B/16.

\vspace{-0.4cm}
\paragraph{Transfer to 11 semantic segmentation benchmarks.}
In Tab.~\ref{tab:compare-ssl}, we report the performance of different SSL pretraining strategies after end-to-end finetuning on semantic segmentation on diverse datasets.
We observe that representations learned with \OURS transfer very well to semantic segmentation.
Of particular interest, MAE representations achieve the second best SSL performance.
In terms of training efficiency, based on our implementation, one \OURS epoch takes 17.4 minutes while one MAE epoch takes 5.7 minutes.
However, \OURS reaches $82.1\%$ average relative improvement over random initialization in 600 epochs while MAE reaches $77.8\%$ in 2.6$\times$ more epochs~(1600).
Hence, \OURS achieves an improvement of $+4.3$ points over MAE while being only 1.1$\times$ longer to pretrain.

\vspace{-0.4cm}
\paragraph{Label-efficient semantic segmentation.}
A good property for pretrained representations is the ability to transfer with few annotations~\cite{alayrac2022flamingo,assran2022masked,zhai2019large}.
In Tab.~\ref{tab:fewshot} we evaluate features when finetuning on fewshot semantic segmentation.
In particular, we follow~\cite{hu2021semi} and randomly sample a fraction of training images from ADE20k and use only those to finetune our model.
In the $1/32$ split, as few as $630$ training images are used.
We report the average over $5$ different folds~\cite{hu2021semi}.
We observe that our spatially-aware pretraining improves label-efficiency of semantic segmentation models.
The gap with previous methods becomes larger when very few images are available for finetuning.

\subsection{Comparison with other pretraining paradigms}
\vspace{-0.1cm}
In this section, we compare our self-supervised location-aware pretraining to two powerful image-level pretraining paradigms:
(i) image classification (i.e. label supervision) as in~\cite{zhai2022scaling,steiner2021train} and
(ii) image-text alignment as in CLIP~\cite{radford2021learning}.

\begin{table}[t]
    \caption{
      \textbf{Comparison with supervised pretrainings} by disentangling localization and classification on semantic segmentation.
     We report classification only (``Classif.'': mAP), localization only (``Loc'': mIoU) and full semantic segmentation (``Both'': mIoU) on ADE20k.
\OURS yields excellent locality and good semantic understanding.
It is behind supervised image-level pretraining on the pure semantic axis (classification) but better on segmentation (``Both'').
}
\centering
\vspace{-0.15cm}
\small
  \setlength{\tabcolsep}{3pt}
    \begin{tabular}{@{} l c c c c c c @{}}
      \toprule
	    Method    & Data & Sup. & Loc-aware? & Classif. & Loc. & Both\\
      \midrule
    \multicolumn{3}{l}{\textit{ViT-Base/16}}\\
      CLIP~\cite{radford2021learning} & WIT & Text & & 58.3 & 66.4 & 45.9 \\
      AugReg~\cite{steiner2021train} & Im21k & Labels & & \textbf{60.7} & 67.4 & 48.1 \\
      \rowcolor{Light}
      \OURS (Ours) & Im21k & $\varnothing$ & \checkmark & 50.2 & \textbf{68.5} & \textbf{48.5}  \\
      \midrule
    \multicolumn{3}{l}{\textit{ViT-Large/16}}\\
       AugReg~\cite{steiner2021train} & Im21k & Labels & & \textbf{60.3} & 68.0 & 50.7 \\
      \rowcolor{Light}
      \OURS (Ours) & Im21k & $\varnothing$ & \checkmark & 51.6 & \textbf{71.0} & \textbf{52.3}  \\
      \bottomrule
 \end{tabular}
    \label{tab:clip_and_sup}
    \vspace{-0.4cm}
\end{table}

\vspace{-0.4cm}
\paragraph{Localization and classification trade-off.}
Semantic segmentation is the coupling of classification and localization, where these two tasks can have different feature preferences.
In Tab~\ref{tab:clip_and_sup}, we disentangle classification and localization performance for models pretrained with an image-level versus spatially-aware objective.
We evaluate performance on classification only by finetuning with a multi-label classification loss.
We evaluate localization only by reporting the performance when replacing the label of each mask by the label of the ground truth mask the model has the best IoU with.
This allows to assess the shape and localization of the predictions but not their class.
We report results for ADE20k in Tab.~\ref{tab:clip_and_sup} and for other datasets in Table~\ref{aptab:sup} in Appendix~\ref{ap2}.
We observe that models pretrained with a global, image-level supervised objective are better than \OURS at classification.
However, \OURS performs better at localization which results in improved performance on semantic segmentation which requires both locality and class-level understanding.

\vspace{-0.4cm}
\paragraph{Depth estimation.}
The previous experiment (Tab.~\ref{tab:clip_and_sup}) shows that \OURS features are particularly good at localization.
While the focus of this work is semantic segmentation, we explore the potential of \OURS on depth estimation, another per-pixel prediction task requiring high spatial understanding but less semantics than semantic segmentation.
We follow~\cite{dehghani2023scaling} and train a Dense Prediction Transformer~\cite{ranftl2021vision} with frozen backbone on the Waymo Open real-world driving dataset~\cite{sun2020scalability}.
We observe in Tab.~\ref{tab:depth_estimation} that \OURS transfers better to depth estimation than backbones trained with image-level supervision.
Notably, \OURS achieves comparable or better performance than supervised ViT-e while using more than 10$\times$ less parameters.

\subsection{Scaling data and model axes}
\label{sec:scaling}
\begin{figure}[t]
\begin{minipage}{0.49\linewidth}
\centering
\includegraphics[width=\linewidth]{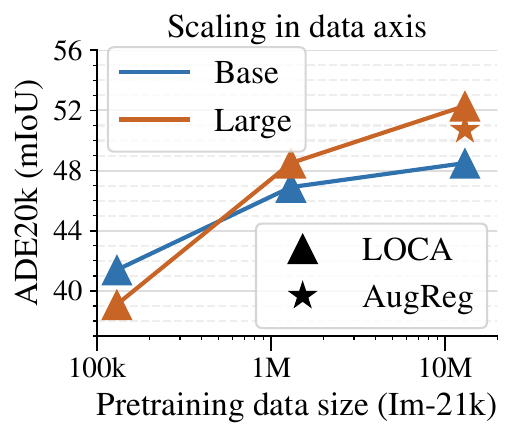}
\end{minipage}
  \begin{minipage}{0.49\linewidth}
  \centering
  \includegraphics[width=\linewidth]{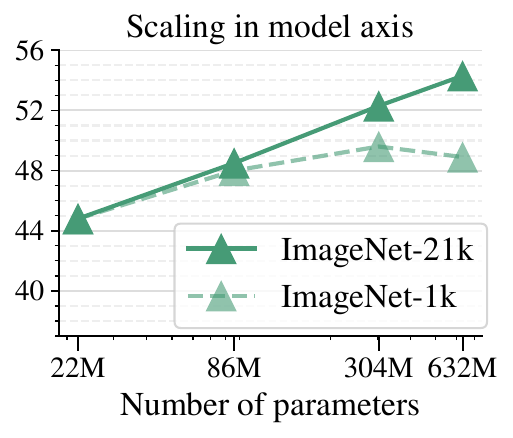}
  \end{minipage}
	\caption{\textbf{Scaling study.}
We report transfer on ADE20k validation set.
Scaling both dataset size and model capacity results in the best of transfer performance.
\label{fig:scaling}
\vspace{-0.6cm}
}
\end{figure}

\vspace{-0.1cm}
A premise of SSL is that it can scale to arbitrary large datasets since images don't require any annotations.
Because location-aware \textit{supervised} pretraining is not feasible in practice due to the huge cost of pixel-level annotations, we believe our self-supervised spatial pretraining could be a good candidate for scaling.
In Fig~\ref{fig:scaling}, we propose a scaling study on data (left panel) and model (right panel) axes.
We observe that \OURS Large architecture benefits more from scaling in dataset size than the smaller Base architecture.
Also, we see that pretraining \OURS on the full ImageNet-21k scales better in model axis than using the smaller, albeit highly curated, ImageNet-1k dataset.
This is not the case for some previous self-supervised learning methods as recently observed by~\cite{skanda2022where}.
Overall, mirroring the trend of image-level supervised pretrainings~\cite{zhai2022scaling,chen2022pali}, we observe that we need to scale both dataset size and model capacity to achieve the best of performance.
While these preliminary results give promising signal about \OURS scalability, we note that ImageNet-21k is a relatively curated dataset and we would still need to probe our model on large, \textit{uncurated} data~\cite{skanda2022where,caron2019unsupervised,tian2021divide,goyal2021self}.

\begin{table}[t]
\caption{\textbf{Monocular depth estimation} on the Waymo Open dataset~\cite{sun2020scalability}.
We follow the setup from~\cite{dehghani2023scaling} and report their number for ViT-L and ViT-e supervised (``sup'') backbones.}
\centering
\vspace{-0.15cm}
\small
\setlength{\tabcolsep}{1pt}
\begin{tabular}{@{} lc ccccc  @{}}
  \toprule
   & & & & \multicolumn{3}{c}{$\delta$ $\uparrow$} \\
   \cmidrule(l{2pt}r{2pt}){5-7}
  Model  & param(M) & MSE $\downarrow$ & AbsRel $\downarrow$ & $< 1.1$& $< 1.25$& $< 1.25^2$\\
  \midrule
 ViT-L sup~\cite{steiner2021train} & 304 & 0.027 & 0.121 & 0.594 & 0.871 & 0.972 \\ %
  \rowcolor{Light}
  ViT-L \OURS & 304 & \textbf{0.024} & \textbf{0.102} & \textbf{0.681} & \textbf{0.891} & \textbf{0.973} \\ %
 \midrule
 ViT-e sup~\cite{zhai2022scaling} & 3926 & \textbf{0.024} & 0.112 & 0.631 & 0.888 & \textbf{0.975} \\ %
 \rowcolor{Light}
 ViT-H \OURS & 632 & \textbf{0.024} & \textbf{0.101} & \textbf{0.685} & \textbf{0.894} & \textbf{0.975} \\ %
  \bottomrule
\end{tabular}
\vspace{-0.4cm}
\label{tab:depth_estimation}
\end{table}

\section{Design Choices Analyses}
\vspace{-0.1cm}
\label{sec:design_choices}
In this section, we detail various design choices for \OURS.
First, we make an in-depth study of the position prediction.
Second, we present an ablation study focused on the pseudo-labeling clustering technique.

\subsection{Position prediction framework}
\vspace{-0.1cm}
\label{sec:localization}
To encourage the network to learn about the spatial arrangement of different object parts, we propose to predict relative positions.
We detail here different components of our framework: the query-reference mechanism, the effect of masking reference patches and the loss function.
Unless specified otherwise, models are trained solely with loss~(\ref{eq:position}) in this section to isolate the effect of position-based training.

\begin{figure}[t]
\includegraphics[width=\linewidth]{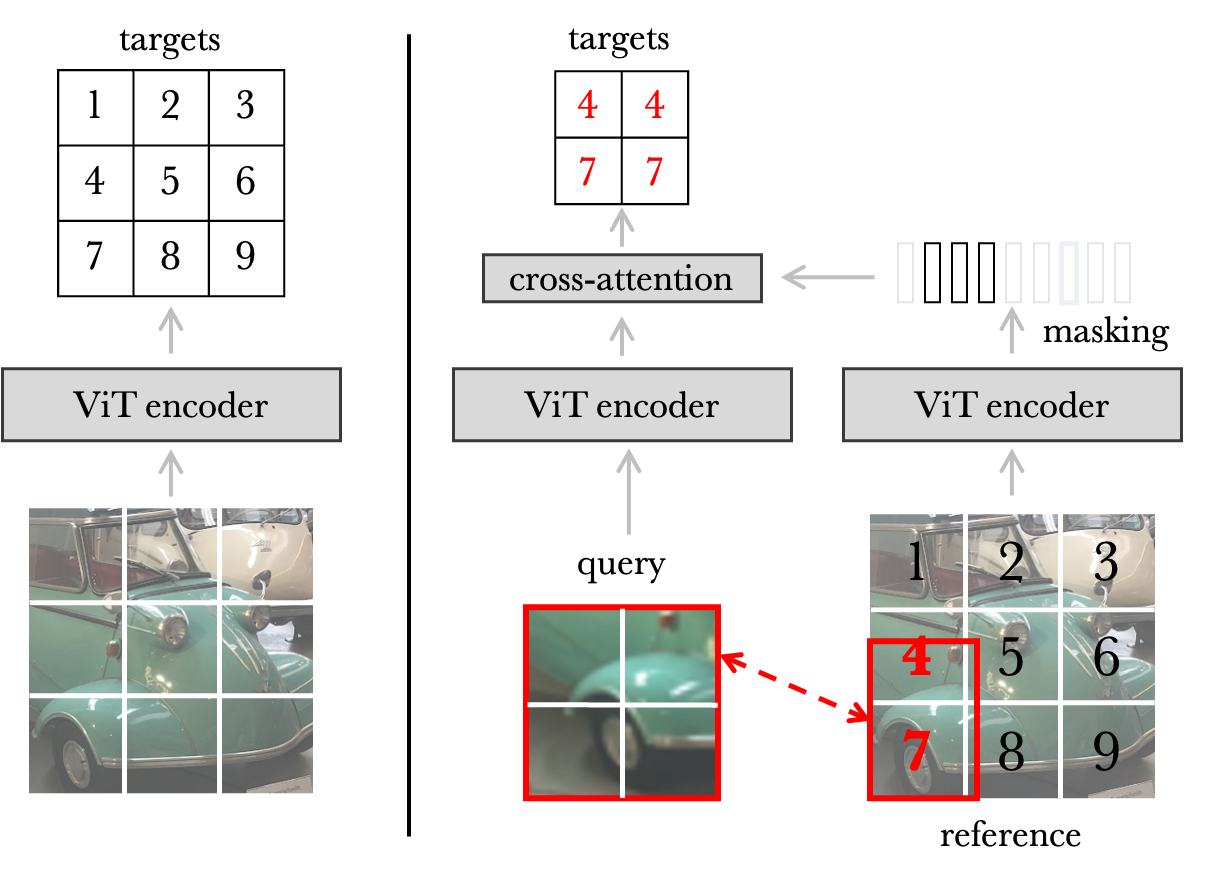}
\vspace{-0.2cm}
\begin{tabular}{cc}
~~~~~~\small{(a) Single}~~~~~~~~~~~~~~~~~~& \small{(b) \OURS query-reference (Ours)}
\end{tabular}
\vspace{-0.1cm}
\caption{\textbf{Conceptual comparison of single vs query-reference} patch position prediction mechanisms:
(a) in a single view as in Zhai~\etal~\cite{zhai2022position};
(b) in a query view relatively to a reference view as in \OURS (Ours).
Quantitative comparison is in Fig.~\ref{fig:patch_pos}.
Masking not illustrated for single.
\vspace{-0.6cm}
\label{fig:query}
}
\end{figure}

\begin{figure}[t]
\includegraphics[width=\linewidth]{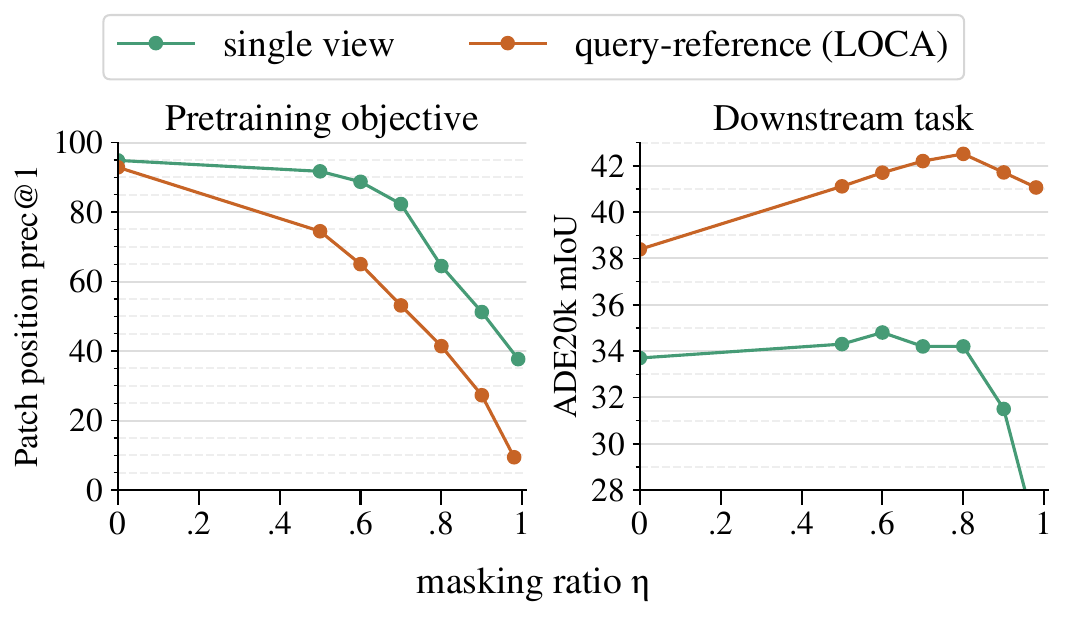}
\vspace{-0.6cm}
\caption{\textbf{Single vs query-reference} patch position prediction mechanisms.
For both mechanisms, we report the position prediction accuracy (left) and the performance after transfer to semantic segmentation on ADE20k (right) for different patch masking ratios $\eta$.
Query-reference makes for a more challenging pre-training objective (lower accuracy on the position prediction task) due to different image statistics between query and reference and constrained patch interactions.
Conceptual differences are illustrated in Fig.~\ref{fig:query}.
Varying the masking ratio controls the difficulty of the task and improves transfer performance.
\label{fig:patch_pos}
\vspace{-0.2cm}
}
\end{figure}

\vspace{-0.4cm}
\paragraph{Query-reference.}
\label{par:query}
We compare the two mechanisms illustrated in Fig.~\ref{fig:query}.
The ``single'' strategy is akin to~Zhai~\etal~\cite{zhai2022position}.
Because position prediction is trivial in single view with positional embeddings~\cite{zhai2022position}, we remove them in this experiment to avoid confounding factors.
We vary $\eta$ the proportion of masked patch tokens.
In ``single'', masking patch tokens means that patches can only attend to the unmasked ones, i.e. only the unmasked patches take part in the computation of attention keys and values~\cite{zhai2022position}.
Results are in Fig.~\ref{fig:patch_pos}.
On the left, we report the validation accuracy for the position prediction task.
This measures how well the network solves its pretraining objective in a fixed training budget, which enables the comparison of the difficulty of different pretraining strategies.
On the right, we show transfer performance.

We see in Fig.~\ref{fig:patch_pos} that the query-reference mechanism of \OURS is a more challenging pretraining framework than Zhai~\etal~\cite{zhai2022position} and leads to better representations for semantic segmentation ($+ 7.6$ mIoU).
This can intuitively be explained by several conceptual differences.
First, in~\cite{zhai2022position}, the network can almost perfectly solve the task by leveraging low-level non-semantic cues such as chromatic aberration, color or edges consistency between patches.
This is \textit{partly} prevented in the query-reference mechanism due to different image statistics between query and reference (thanks to cropping, rescaling and color jittering).
Second, the way query (i.e. patches that predict a position) and reference (i.e. context patches) can interact is in stark contrast in the two mechanisms.
In ``single'', query and reference interact in an unconstrained manner at all stages of the computation.
With masking, this design is party modified by processing each query patch independently but still allowing them to fully attend to the reference patches at each block.
By contrast, in \OURS, query patches can attend freely to each other but cannot look at the reference patches until the last stage of the network.
Intuitively, this more constrained interaction encourages both query and reference patches to develop stronger final localization features.

\begin{figure}[t]
\begin{minipage}{0.5\linewidth}
\centering
\includegraphics[width=\linewidth]{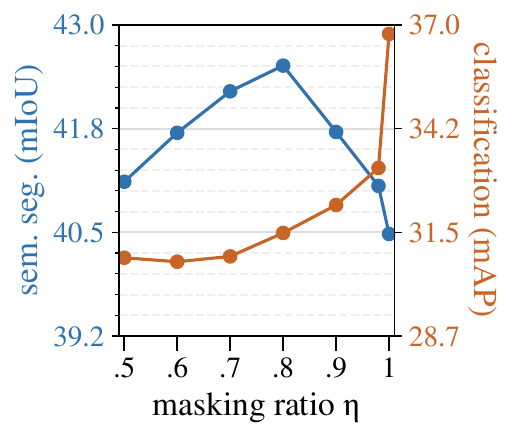}
\vspace{-1.2cm}
\end{minipage}
  \begin{minipage}{0.45\linewidth}
	\caption{\textbf{Masking reference patches to the query} improves both classification and segmentation on ADE20k.
Too much masking prevents the query from solving the task by spatial reasoning which hurts segmentation.
  \vspace{-1.2cm}
	}
\label{fig:ratio}
\end{minipage}
\end{figure}

\begin{figure}[t]
\centering
\renewcommand{\arraystretch}{0}
\begin{tabular}{cccc}
&~~~~~~\footnotesize{$\eta = 0$}&~~~~~~~~~\footnotesize{$\eta = 0.8$}&\footnotesize{$\eta = 1$} \\
\multicolumn{4}{c}{\includegraphics[width=0.95\linewidth]{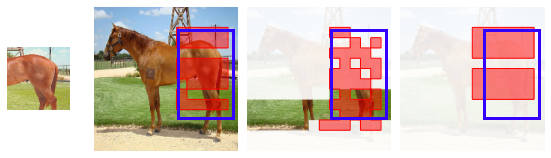}}
\\
\multicolumn{4}{c}{\includegraphics[width=0.95\linewidth]{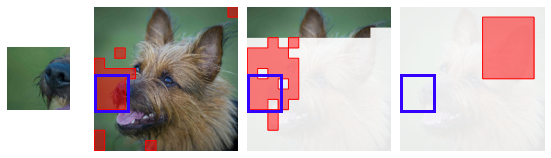}}
\\
\multicolumn{4}{c}{\includegraphics[width=0.95\linewidth]{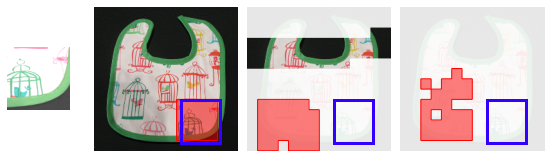}}
\\
\multicolumn{4}{c}{\includegraphics[width=0.95\linewidth]{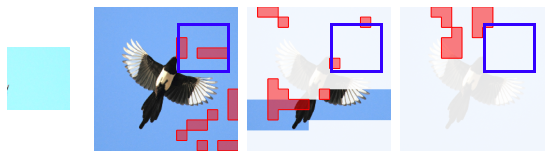}}
\\
~~\footnotesize{Query} & \multicolumn{3}{c}{\footnotesize{Reference}} \\
\end{tabular}
\caption{
  \textbf{Visualizing LOCA's predictions.}
  The query location is shown in blue in the reference and \OURS predictions are shown in red.
  Columns correspond to different reference masking rates and we show only patches visible to the query.
  More examples in the Appendix~\ref{ap:vis}.
  Displayed images are not seen during training.
  \vspace{-0.6cm}
}
\label{fig:visu}
\end{figure}

\vspace{-0.4cm}
\paragraph{Masking reference patches.}
In Fig.~\ref{fig:patch_pos}, we observe that the localization pretraining task can be solved near perfectly when all the patches in the reference are visible to the query (see Fig.~\ref{fig:patch_pos}~left for $\eta = 0$).
Masking to the query makes the pretraining objective more challenging and leads to better representations.
In Fig.~\ref{fig:ratio}, we analyze this effect further.
We consider different masking ratios and report for the same downstream dataset both the transfer performance on semantic segmentation and multi-label classification (with frozen backbone) by turning the semantic segmentation annotations into classification labels.
We observe in Fig.~\ref{fig:ratio} that masking improves both localization and classification capabilities of the network.
Intuitively this is because masking reference patches forces the query to rely less on finding matching salient points between the two views and more on recognizing objects and their parts as illustrated in Fig~\ref{fig:visu}.

However, when masking is too aggressive, the query does not see enough of the reference to solve its task by relative localization and resorts to other cues.
To understand this phenomenon, we push masking to extreme rates and even report performance when the reference is \textit{not visible at all} ($\eta = 1$).
Surprisingly, we find that the query still manages to solve the localization pretraining task to some extent with a localization accuracy of 3.7\% (random guessing achieves 0.5\%).
We hypothesize that two ways of solving the task without looking at the reference are to (i) learn where things are typically located in images and (ii) memorize all the dataset images.
We argue that the ``memorization'' regime is akin to an implicit formulation of the ``exemplar'' instance discrimination approach of Dosovitskiy~\etal~\cite{dosovitskiy2016discriminative} where the network learns to recognize each individual instance of a dataset (but without a classifier of the size of the dataset as in~\cite{dosovitskiy2016discriminative}).
Overall, both learning biases of general dataset statistics and instance discrimination have been shown to improve transfer performance on classification downstream tasks~\cite{dosovitskiy2016discriminative,wu2018unsupervised,gidaris2018unsupervised} which is consistent with the boost in classification observed for $\eta=1$.

Finally, this experiment shows that an optimal masking ratio for semantic segmentation features is high, but not too high either so that the network can still solve the task by \textit{relative localization}.
In practice, we use $\eta=0.8$.

\vspace{-0.4cm}
\paragraph{Choice of localization loss.}
First, we compare predicting the position of all patches versus the position of the central patch only.
We see in Tab.~\ref{tab:loss} that all patches is better.
We hypothesize that this is because it requires to predict the spatial extent of the query and not just an anchor point.
Second, we compare solving a per-patch position classification problem versus regressing the coordinates of the query box in the reference.
For box prediction, we use a linear combination of $\ell_1$ loss and the generalized IoU loss, following UP-DETR~\cite{dai2021up,carion2020end}.
Because query and reference patchification grids are usually not aligned, matching patches in query and reference do not have exactly the same content.
This does not affect the box regression formulation, which might give it an advantage over per-patch classification.
However, we surprisingly find in Tab.~\ref{tab:loss} that box regression leads to poorer performance than per-patch classification.
It is possible that this loss requires additional hyper-parameter tuning (we use the default in~\cite{carion2020end,dai2021up}).
Overall, position classification is a simple implementation of the relative localization problem and works well in practice.
\vspace{-0.4cm}
\paragraph{Combining with patch clustering.}
In the previous experiments, we have validated our position prediction scheme and showed that it improves by $+7.6$mIoU over the position prediction method of~Zhai~\etal~\cite{zhai2022position}.
While we find that predicting position only is performing less well than predicting patch-level cluster assignments only ($-3.3$mIoU) the best performance is obtained when predicting \textit{both} ($+0.7$mIoU over cluster only) which demonstrates some complementary between them.

\begin{table}[t!]
    \caption{
      \textbf{Localization loss.}
We report mIoU on ADE20k for different loss variants.
Predicting the position of all patches \textit{vs} the position of the central patch only is better, likely because it involves reasoning about the spatial extent of the query.
Classification works better than regression in our experiment, despite the fact that regression is not impacted by the misalignment between query and reference patch grids.
}
\vspace{-0.15cm}
\centering
\small
  \setlength{\tabcolsep}{2pt}
    \begin{tabular}{@{} l cc c@{}}
      \toprule
	   Output & Predicts spatial extent & Loss & ADE20k \\
      \midrule
\rowcolor{Light}
Every patch position & \checkmark & Classif. & 42.5 \\
Central patch position & & Classif. & 38.6 \\
Query box coordinates & \checkmark & Regress. & 39.0 \\
      \bottomrule
 \end{tabular}
    \label{tab:loss}
    \vspace{-0.2cm}
\end{table}

\vspace{-0.4cm}
\paragraph{Visualizing \OURS's predictions.}
In Fig.~\ref{fig:visu}, we visualize the output of location prediction models trained with different masking rates: $\eta=0$ (no masking), $\eta=0.8$ (default) and $\eta=1$ (invisible reference).
The first row shows a situation where the network can make a valid guess about the query's location solely based on the query visual appearance, i.e. without looking at the reference.
In the second row, we see that \OURS successfully locates the snout of the dog based on the reference ear patches.
This suggests that it has learned about spatial arrangement of different parts of a dog.
Third row depicts the case where the network can leverage low-level cues such as edge consistency to locate the query.
The masked variants are restrained in their use of such cues and hence fail to locate the query.
Finally, in last row, there is no visible cue in the query that allows its localization.
The prediction is degenerated for all the variants.

\subsection{Ablation study of patch clustering}
\vspace{-0.1cm}
In Tab.~\ref{fig:multi_ablat}, we report model ablation results.
In Tab.~\ref{fig:multi_ablat}~a), we observe that both Sinkhorn-Knopp and me-max regularizations are useful to encourage the model to use the full set of cluster prototypes.
In Tab.~\ref{fig:multi_ablat}~b), we compare our patch-level pseudo-labeling method to image-level ones on ADE20k semantic segmentation and on ImageNet-1k 10-shot classification.
The image-level clustering framework is akin to existing SSL frameworks such as DINO~\cite{caron2021emerging} or MSN~\cite{assran2022masked}.
We evaluate two global aggregation techniques: token (``CLS'') and global average pooling (GAP).
We observe that performance on semantic segmentation is improved when considering per-patch assignments instead of image-level clustering.
However, we observe a decay on classification.
In Tab.~\ref{fig:multi_ablat}~c), we see that the method is robust to the number of clusters $K$, though over-clustering is beneficial~\cite{ziegler2022self,caron2018deep}.
In  Tab.~\ref{fig:multi_ablat}~d), we show the effect of reducing the number of queries.
Using a single query instead of 10 allows to speed up pretraining time by $\times 3$ but induces a loss of $5.1$mIoU in transfer performance.

\begin{figure}[t]
\begin{minipage}{0.45\linewidth}
\centering
\small
  \setlength{\tabcolsep}{2pt}
   \caption*{\small{a) SK and me-max encourage use of all prototypes. H:~average prediction entropy.}}
    \begin{tabular}{@{} c c c c @{}}
      \toprule
	    SK & me-max & H & ADE20k \\
      \midrule
      \rowcolor{Light}
      \checkmark & \checkmark & 8.28 & 46.2 \\ 
      \checkmark &  & 8.27 & 46.1 \\
       & \checkmark & 8.14 & 45.7 \\
       & & 0 & collapse \\
      \bottomrule
 \end{tabular}
   \setlength{\tabcolsep}{1pt}
      \caption*{\small{c) Effect of the number of cluster prototypes.}}
    \begin{tabular}{@{}l c c c c@{}}
      \toprule
	   $K$ & $2^{4}$ & $2^{8}$ & \colorbox{Light}{~$2^{12}$} & $2^{14}$ \\
      \midrule
      ADE20k & 45.3 & 46.2 & \colorbox{Light}{46.2} & 45.8\\
      \bottomrule
 \end{tabular}
\end{minipage}\quad
\begin{minipage}{0.45\linewidth}
\centering
\small
  \setlength{\tabcolsep}{1.pt}
   \caption*{\small{b) Patch-level clustering is better for transfer on semantic segmentation.}}
    \begin{tabular}{@{} l c c @{}}
   
      \toprule
	     & classif. & semseg. \\
	    Cluster & Im1k-10s & ADE20k \\
      \midrule
      \rowcolor{Light}
      Patch & 40.9 & 46.2 \\
      Image (CLS) & 48.6 & 43.8 \\
      Image (GAP) & 48.5 & 43.8  \\
      \bottomrule
 \end{tabular}
      \caption*{\small{d) Number of queries.}}
  \setlength{\tabcolsep}{1.pt}
    \begin{tabular}{@{}l c c c c @{}}
      \toprule
	   \# queries & \colorbox{Light}{~10~~} & 5 & 2 & 1 \\
      \midrule
      ADE20k & \colorbox{Light}{46.2} & 45.5 & 43.4 & 41.1 \\
      Speedup & -- & $\times$1.5 & $\times$2.1 & $\times$3.0 \\
      \bottomrule
 \end{tabular}
\end{minipage}
\captionof{table}{\textbf{Ablation study of different design choices.}}
\vspace{-0.5cm}
\label{fig:multi_ablat}
\end{figure}

\section{Conclusion}
We propose a SSL spatially-aware pretraining that transfers well to semantic segmentation.
A promising direction for future work is to combine dense objectives with global image-level ones~\cite{zhou2021ibot,bardes2022vicreg}.
Finally, we focus on semantic segmentation in this paper as it is representative of tasks requiring spatial reasoning.
However, the set of visual task with localization is large and we hope that our findings can serve as a useful checkpoint for future studies beyond the scopes of semantic segmentation or even still images.

\paragraph{Acknowledgement.}
We thank Anurag Arnab, Lucas Beyer, Mostafa Dehghani, Ahmet Iscen, Thomas Kipf, Thomas Lucas, Thomas Mensink, Matthias Minderer, Basil Mustafa as well as the entire Ganesha team and Grand Vision group for their precious help, support and discussions.

{\small
\bibliographystyle{ieee_fullname}
\bibliography{egbib}
}
\newpage
\renewcommand{\thesubsection}{\Alph{subsection}}
\newcommand\tab[1][5mm]{\hspace*{#1}}
\section*{Appendix}

\begin{table}[t]
    \caption{
      \textbf{Comparison with supervised pretrainings} by disentangling localization and classification on semantic segmentation.
     We report classification only with a frozen backbone (``Classif.'': mAP), localization only (``Loc'': mIoU) and semantic segmentation end-to-end finetunings (``Both'': mIoU) on ADE20k (``A'') and Pascal Context (``P'').
     Results for ADE20k are also presented in the main paper.
\OURS yields excellent locality and good semantic understanding.
It is behind supervised image-level pretraining on the pure semantic axis (classification) but better on segmentation (``Both'').
}
\centering
\vspace{-0.15cm}
\small
  \setlength{\tabcolsep}{2pt}
    \begin{tabular}{@{} l c c c cc c cc c cc @{}}
      \toprule
      
	       & & && \multicolumn{2}{c}{Classif.} && \multicolumn{2}{c}{Loc.} && \multicolumn{2}{c}{Both} \\
	    \cmidrule{5-6}\cmidrule{8-9}\cmidrule{11-12}
	   Method    & Data & Sup. && A & P && A & P && A & P \\
      \midrule
    \multicolumn{3}{l}{\textit{ViT-Base/16}}\\
      CLIP~\cite{radford2021learning} & WIT & Text && 58.3 & 67.1 && 66.4 & 73.2 && 45.9 & 52.8 \\
      AugReg~\cite{steiner2021train} & Im21k & Labels&& \textbf{60.7} & \textbf{66.1} && 67.4 & 75.0 && 48.1 & \textbf{55.7} \\
      \rowcolor{Light}
      \OURS & Im21k & $\varnothing$  && 50.2 & 63.9 && \textbf{68.5} & \textbf{76.5} && \textbf{48.5} & \textbf{55.7} \\
      \midrule
    \multicolumn{3}{l}{\textit{ViT-Large/16}}\\
       AugReg~\cite{steiner2021train} & Im21k & Labels && \textbf{60.3} & \textbf{65.8} && 68.0 & 75.4 && 50.7 & 56.5  \\
      \rowcolor{Light}
      \OURS & Im21k & $\varnothing$ && 51.6 & 63.3 && \textbf{71.0} & \textbf{78.9} && \textbf{52.3} & \textbf{60.3}  \\
      \bottomrule
 \end{tabular}
    \label{aptab:sup}
    \vspace{-0.4cm}
\end{table}

\begin{table*}[t!]
    \caption{
      \textbf{Comparison with previous results on 11 semantic segmentation datasets.} 
We report mean IoU on the validation set of different semantic segmentation benchmarks.
Backbones are pretrained using different self-supervised and supervised methods.
We consider two settings:
(i) pretraining on ImageNet-1k with ViT-Base/16 and (ii) pretraining on ImageNet-21k with ViT-Large/16.
We follow the experimental setup of Segmenter~\cite{strudel2021segmenter} for end-to-end finetuning with linear decoder.
We report official numbers from~\cite{strudel2021segmenter} when available and run the evaluation from official released checkpoints when not available.
We report the average over $5$ runs with single-scale mode (*: with multi-scale evaluation).
Finally, we report in the last column the relative improvement over starting from random initialization averaged over the 11 datasets (``avg.rel $\Delta$'').
}
\centering
\small
  \setlength{\tabcolsep}{2.3pt}
      \begin{tabular}{@{} l c ccc c ccccc c c c c c c | c@{}}
      \toprule
      & & \multicolumn{3}{c}{\textit{Consumer}} && \multicolumn{5}{c}{\textit{Driving}} && \multicolumn{1}{c}{\textit{Indoor}} && \multicolumn{1}{c}{\textit{Aerial}} && \multicolumn{1}{c|}{\textit{Underwater}} & \multicolumn{1}{c}{Avg. rel. } \\
\cmidrule{3-5}\cmidrule{7-11}\cmidrule{13-13}\cmidrule{15-15}\cmidrule{17-17}
	    Pretraining method   &  Labels & ADE20k & P.Cont & P.VOC && Citys. & BDD & CamVid & IDD & KITTI && SUN && ISPRS && SUIM & $\Delta$ (\%)\\
      \midrule
\multicolumn{4}{l}{\textit{ImageNet-1k / ViT-Base/16}} & \multicolumn{13}{c|}{}\\
\gray{Random init.} & & \gray{21.1} & \gray{19.6} & \gray{29.1} && \gray{51.4} &  \gray{40.2} & \gray{43.3} & \gray{45.2} & \gray{39.0} && \gray{19.7} && \gray{28.1} && \gray{53.0} & \gray{0} \\
DeiT~\cite{touvron2020training,strudel2021segmenter} &  \checkmark & 47.1 & -- & -- && -- & -- & -- & -- & -- && -- && -- && -- & -- \\
DeiT-III~\cite{touvron2022deit} & \checkmark & 47.3 & 53.9 & 76.1 && 79.7 & 62.7 & 53.8 & 55.4 & 47.2 && 47.5 && 42.1 && 73.5 & +79.0 \\
DINO~\cite{caron2021emerging} & & 44.1 & 50.7 & 74.1 && 78.4 & 60.7 & 51.5 & 54.3 & 46.4 && 44.4 && 41.5 && 71.2 & +71.9\\
MoCo-v3~\cite{chen2021empirical} & & 45.4 & 51.6 & 74.5 && 78.6 & 60.4 & 51.1 & 53.7 & 45.7 && 45.6 && 42.1 && 72.6 & +73.6 \\
iBOT~\cite{zhou2021ibot} & & 47.0 & 54.6 & 75.0 && \textbf{79.8} & 62.1 & 51.5 & 55.5 & 47.0 && 46.3 && 42.2 && 73.2 & +77.7 \\
MAE~\cite{he2022masked} & & 45.5 & 51.7 & 75.0 && 79.7 & 62.1 & \textbf{57.8} & \textbf{55.8} & 48.3 && 45.9 && 44.6 && 72.4 & +77.8 \\
\rowcolor{Light}
\OURS (Ours) & & \textbf{47.9} & \textbf{54.9} & \textbf{76.7} && \textbf{79.8} & \textbf{62.8} & 56.1 & 55.6 & \textbf{48.5} && \textbf{47.7} && \textbf{45.6} && \textbf{74.0} & \textbf{+82.1} \\
\midrule
\midrule
\multicolumn{4}{l}{\textit{ImageNet-21k / ViT-Large/16}}& \multicolumn{13}{c|}{}\\
\gray{Random init.} & & \gray{21.2} & \gray{20.1} & \gray{31.1} && \gray{44.9} &  \gray{39.7} & \gray{43.7} & \gray{45.4} & \gray{39.7} && \gray{19.2} && \gray{26.7} && \gray{48.3} & \gray{0} \\
Augreg~\cite{steiner2021train,strudel2021segmenter} &  \checkmark & 50.7  & ~~56.5* & 77.5 && ~~80.7* & 62.3 & 51.2 & 54.9 & 47.6 && 48.5 && 43.8 && 73.7 & +84.8 \\ 
\rowcolor{Light}
\OURS (Ours) &  & \textbf{52.3} & \textbf{60.3} & \textbf{78.7} && \textbf{81.5} & \textbf{65.3} &  \textbf{56.0} &  \textbf{57.5} & \textbf{50.3} && \textbf{51.3} && \textbf{49.7} && \textbf{73.7} &  \textbf{+93.9}\\ 
\bottomrule
 \end{tabular}
    \label{aptab:sota}
\end{table*}

\subsection{Implementation and Evaluation Details} \label{ap1}
\subsubsection{\OURS pretraining details}
We train our models with a base learning rate of $0.001$ (linearly ramped up during the first $15$ epochs before cosine decay), a batch size of $1024$ and a weight decay of $0.1$ with adamw optimizer~\cite{loshchilov2018fixing}.
Models for ablations and analyses are trained during 100 epochs while checkpoints for main results are trained for 600 epochs.
100 epochs of training on 16 TPUv2 accelerators take 29 hours.
We use $\eta = 0.8$ for masking.
For data augmentation we apply random resized crop, horizontal flipping and color jittering (following the parameters from BYOL~\cite{grill2020bootstrap}).
Momentum parameter is set to $0.996$ and increased with a cosine schedule to $1$ during training~\cite{grill2020bootstrap,zhou2021ibot,caron2021emerging}.
We typically use $10$ queries per reference view.
We follow MSN pipeline for generating query views~\cite{assran2022masked}.
In particular, we restrain the spatial extent of the queries thanks to token dropping.
Specifically, one query undergoes random token dropping while the other queries have focal random token dropping.
Results are reported with the weights from the momentum branch~\cite{caron2021emerging,zhou2021ibot}.
We implement \OURS in Jax using the open-sourced \textsc{scenic} library~\cite{dehghani2021scenic}.
Code and models to reproduce our results will be made publicly available as a \textsc{scenic} project.

\subsubsection{Eleven semantic segmentation datasets} \label{ap:transfer}
In this paper, we report results on the following diverse semantic segmentation benchmarks:
ADE20k~\cite{zhou2017scene}, Pascal Context~(``P.Cont'')~\cite{mottaghi2014role}, Pascal VOC~(``P.VOC'')~\cite{everingham2010pascal}, Cityscapes~(``Citys.'')~\cite{cordts2016cityscapes}, Berkeley Deep Drive (``BDD'')~\cite{yu2020bdd100k}, CamVid~\cite{camvid}, India Driving Dataset (``IDD'')~\cite{varma2019idd}, KITTI~\cite{abu2018augmented}, SUN-RGB-D~(``SUN'')~\cite{song2015sun}, ISPRS~\cite{meidow2014theme} and SUIM~\cite{islam2020semantic}.
We detail the main four datasets used in this paper here and refer to corresponding papers and to Mensink~\etal~\cite{mensink2021factors} for details on the remaining other datasets.\\

\noindent \textbf{ADE20K~\cite{zhou2017scene}.} It is a dataset containing scenes with fine-grained labels with 150 semantic classes and is one of the most challenging semantic segmentation datasets.
The training split is composed of 20,210 images.
We report results on the validation set, composed of 2,000 images.

\noindent \textbf{Pascal Context~\cite{mottaghi2014role}.}
The training split is composed of 4,998 images with 59 semantic classes and a background class (hence a total of 60 classes).
The validation set has 5,105 images.

\noindent \textbf{Pascal VOC~\cite{everingham2010pascal}.}
This dataset has a training set of 10,582 images and counts 21 classes (with background class).
We report results on the validation set, it has 1,449 images.

\noindent \textbf{Cityscapes~\cite{cordts2016cityscapes}.} The dataset contains 5,000 images from 50 different cities. We consider the setup with 19 classes as in~\cite{strudel2021segmenter}. There are 2,975 images in the training set, 500 images in the validation set and 1,525 images in the test set (not used). We report results on the validation set.

\subsubsection{Evaluation protocol} \label{ap:transfer2}
We hope to use a simple decoder for semantic segmentation for better investigating the effectiveness of pretraining.
We precisely follow the experimental setup of Segmenter~\cite{strudel2021segmenter} for end-to-end finetuning of Vision Transformer with linear decoder.
The data augmentation used during training is normalization, random resizing of the image to a ratio between 0.5 and 2.0, photometric jittering and random horizontal flipping.
We randomly crop images and use padding to preserve aspect ratio.
We use the $512 \times 512$ resolution for all datasets and $768 \times 768$ on Cityscapes.
On ADE20k, we train for 127 epochs with minibatch size of 16 (resulting in 160k iterations).
On Pascal, we train for 256 epochs with minibatch size of 16 (resulting in 80k iterations).
On Cityscapes, we train for 215 epochs with minibatch size of 8 (resulting in 80k iterations).
On all other datasets, we train with minibatch size of 16 and 160k iterations.
We use the ``poly'' learning rate decay schedule and sweep the base learning rate in $\{8e-5, 1e-4, 3e-4, 8e-4\}$ for all of our runs.
Weight-decay is kept fixed at $0.01$.
At evaluation time, we use the sliding-window mechanism with window resolution matching the resolution used during training (i.e. $512 \times 512$ for all datasets and $768 \times 768$ for Cityscapes) to handle varying image sizes during inference.
Table 3 row 6 in Segmenter paper~\cite{strudel2021segmenter} reports $48.06$ mIoU (single scale) for finetuning from ViT-B/16 AugReg checkpoint~\cite{steiner2021train}.
The average of 3 runs in the same setup in our codebase gives $48.07$ mIoU (run 1: $48.41$, run 2: $48.08$, run 3: $47.70$).
This validates our reproduction of the linear decoder presented in the Segmenter work~\cite{strudel2021segmenter}.

\subsection{Additional Results}

\subsubsection{Comparison on 11 semantic segmentation tasks}

In Table~\ref{aptab:sota}, we compare \OURS pre-training to different self-supervised and supervised methods on eleven semantic segmentation benchmarks with diverse properties and domains.
The datasets and evaluation protocols are detailed in Sections~\ref{ap:transfer}~and~\ref{ap:transfer2}.
With ViT-Base/16 architecture and ImageNet-1k dataset, the relative improvement over starting from random initialization averaged over the 11 datasets for \OURS features is $+82.1\%$.
This is $+4.8$ points above the best self-supervised competitor, MAE, and $+3.1$ points above supervised pretraining with DeiT-3.
With ViT-Large/16, \OURS features transfer even better to semantic segmentation.
They reach a relative improvement over random initialization of $+93.9\%$, which is 9.1 points higher than the results obtained with AugReg checkpoint~\cite{steiner2021train} in the Segmenter paper~\cite{strudel2021segmenter}.
This validates our location-aware pretraining for transferring on semantic segmentation downstream tasks compared to using checkpoints pretrained with a supervised, global task such as AugReg~\cite{steiner2021train}.

\begin{table*}[t!]
    \caption{
\textbf{Disentangling localization and classification on semantic segmentation.}
We report end-to-end finetuning on classification only (with a multi-label classification loss) and localization only (with an oracle giving the class of the segmentation masks) evaluations on 4 popular semantic segmentation benchmarks: ADE20k~\cite{zhou2017scene}, Pascal Context (``P.Cont.'')~\cite{mottaghi2014role}, Pascal VOC (``P.VOC'')~\cite{everingham2010pascal} and Cityscapes (``City.'')~\cite{cordts2016cityscapes}.
Best number is in bold and second best is underlined.
We report performance for different methods pretrained on ImageNet-1k (with or without labels) with ViT-B/16.
}
\centering
\small
  \setlength{\tabcolsep}{1.5pt}
    \begin{tabular}{@{} l cccc c cccc c cccc@{}}
      \toprule
      & \multicolumn{4}{c}{Classification only (mAP)} &~~~~& \multicolumn{4}{c}{Localization only (mIoU)} &~~~~& \multicolumn{4}{c}{Both (mIoU)} \\
\cmidrule{2-5}\cmidrule{7-10}\cmidrule{12-15}
	   Method    & ADE20k & P. Cont. & P. VOC & Citysc. && ADE20k & P. Cont. & P. VOC & Citysc. && ADE20k & P. Cont. & P. VOC & Citysc. \\
	   
      \midrule
\textit{Image-level pretrainings} \\
  DINO~\cite{caron2021emerging} & 61.6 & 67.7 & 89.9 & 81.5 && 64.5 & 71.6 & 78.7 & 79.6 && 44.1 & 50.7 & 74.1  & 78.4 \\
  MoCo-v3~\cite{chen2021empirical} & 61.1 & 69.3 & 93.6 & 82.1 && 66.2 & 73.7 & 79.0 & 79.9 && 45.4 & 51.6 & 74.5 & 78.6 \\
  Supervised (DeiT-III~\cite{touvron2022deit}) & \textbf{64.8} & \textbf{71.5} & \textbf{94.6} & \underline{84.0} && 66.5 & 73.6 & \underline{80.1} & 80.7 && \underline{47.3} & \underline{53.9} & \underline{76.1} & \underline{79.7} \\
\midrule
\textit{Spatially-aware pretrainings} \\
  MAE~\cite{he2022masked} & 59.0 & 67.6 & 92.8 & \textbf{84.3} && \underline{67.0} & \underline{74.3} &  79.9 & \underline{81.1} && 45.5 & 51.7 & 75.0 & \underline{79.7 }\\
  \rowcolor{Light}
 \OURS (Ours) & \underline{62.2} & \underline{69.9} & \underline{93.7} & 83.6 && \textbf{67.9} & \textbf{75.4} & \textbf{80.5} & \textbf{81.4} && \textbf{47.9} &\textbf{54.9}  & \textbf{76.7} & \textbf{79.8}  \\
      \bottomrule
 \end{tabular}
    \label{tab:tradeoff}
\end{table*}

\subsubsection{More localization/classification trade-off results} \label{ap2}

Semantic segmentation is the coupling of classification and localization, where these two tasks can have different feature preferences.
In this section, we propose to disentangle classification and localization performance on semantic segmentation benchmarks which require both.
First, we discard local information and evaluate classification only by training a linear layer with a multi-label binary cross-entropy loss.
Second, we evaluate localization only by reporting the performance of an already finetuned semantic segmentation model in presence of a class oracle.
Specifically, the oracle replaces the label of each mask by the label of the ground truth mask it has the best IoU with.
This evaluation allows to assess the shape and localization of the predictions but not their class.

\paragraph{Comparison with image-level supervised pretrainings.}
We compare our self-supervised location-aware pretraining to two powerful image-level pretraining paradigms:
(i) image classification (i.e. label supervision) as in~\cite{zhai2022scaling,steiner2021train} and
(ii) image-text alignment as in CLIP~\cite{radford2021learning}.
We present the results by disentangling localization and classification on semantic segmentation.
Note that we report classification with a frozen backbone as typically done in self-supervised learning literature (coined as the ``linear probing'' evaluation protocol).
In Table 3 of the main paper, we have reported results only with ADE20k dataset.
We show in Table~\ref{aptab:sup} that observations and conclusions are consistent when considering other datasets, namely Pascal Context and Cityscapes.
On Pascal Context, we interestingly observe in~Table~\ref{aptab:sup} that the final performance on semantic segmentation is the same for AugReg and \OURS ViT-B/16 checkpoints pretrained on ImageNet-21k (i.e. 55.7 mIoU).
However, this performance can be explained by different factors for the two checkpoints:
(i) good classification performance for AugReg (i.e. 66.1 for AugReg \textit{vs} 63.9 for \OURS) and (ii) acute localization performance for \OURS (i.e. 75.0 for AugReg \textit{vs} 76.5 for \OURS).

\paragraph{Comparison with different supervised and self-supervised pretrainings.}
In Table~\ref{tab:tradeoff}, we compare the behavior of models pretrained with an image-level versus spatially-aware objective with ViT-B/16 on ImageNet-1k.
Unlike previous experiment in Table~\ref{aptab:sup}, we report end-to-end finetuning for classification only in this experiment.
Indeed, we have observed that freezing the backbone and training a linear classifier on top of MAE features perform very poorly~\cite{he2022masked}.
In Table~\ref{tab:tradeoff}, we observe that models pretrained with a global, image-level objective such as DeiT-III or MoCo-v3 tend to be better on the classification aspect.
By contrast, models trained with a spatially-aware objective such as MAE or \OURS produce more spatially accurate predictions.
Overall, \OURS yields excellent locality and good class-level understanding (while not beating representations learned with label classification pretraining~\cite{touvron2022deit} on the pure classification axis).
This results in strong semantic segmentations which require both locality and semantic features.

\subsubsection{Scaling Study}
We report in Table~\ref{aptab:scaling1} (resp. in Table~\ref{aptab:scaling2}) the numbers corresponding to Figure~2~(left) (resp.~(right)) of the main paper.
We observe that the performance boost from increasing the pretraining dataset size increases when considering bigger architectures.

\begin{table}[t]
    \caption{
      \textbf{Scaling in data axis} on ImageNet-21k. 
We report performance (mean IoU on ADE20k - single scale evaluation) for different pretrained \OURS models.
``Rand-\textit{x}'' means that we take a random subset of size \textit{x} in ImageNet-21k.
`INet-1k'' means that we use the ImageNet-1k dataset only for pretraining.
}
\centering
\small
  \setlength{\tabcolsep}{3.5pt}
    \begin{tabular}{@{} l c c c c @{}}
      \toprule
	    Arch / Data      & Rand-130k & Rand-1.3M & Full 13M & INet-1k \\
      \midrule
     ViT-Base/16 & 41.4 & 46.9 & 48.5 & 47.9 \\
     ViT-Large/16 & 39.1 & 48.5 & 52.3 & 49.6 \\
      \bottomrule
 \end{tabular}
    \label{aptab:scaling1}
\end{table}

\begin{table}[t]
    \caption{
      \textbf{Scaling in model axis} on ImageNet-21 and ImageNet-1k. 
We report performance (mean IoU on ADE20k - single scale evaluation) for different pretrained \OURS models.
The performance boost from increasing the pretraining dataset size increases when considering bigger architectures.
}
\centering
\small
  \setlength{\tabcolsep}{3pt}
    \begin{tabular}{@{} l c c c c @{}}
      \toprule
	    Data / Arch      & Small/16 & Base/16 & Large/16 & Huge/16 \\
      \midrule
     ImageNet-1k & 44.8 & 48.0  & 49.6 & 48.9 \\
     ImageNet-21k & 44.8 \footnotesize{(+0.0)} & 48.5 \footnotesize{(+0.5)} & 52.3 \footnotesize{(+2.7)} & 54.3 \footnotesize{(+5.4)} \\
      \bottomrule
 \end{tabular}
    \label{aptab:scaling2}
\end{table}

\subsection{Visualizations}
\label{ap:vis}
\begin{figure*}[t]
\centering
\begin{tabular}{cccc cccc}
~~~\footnotesize{query}&~~~~~~~~~\footnotesize{$\eta = 0$}&~~~~~~~~~~~~\footnotesize{$\eta = 0.8$}&\footnotesize{$\eta = 1$} &~~\footnotesize{query}~~~~~~~~~~~~~~~~~~~\footnotesize{$\eta = 0$}&~~~~~~~~~~~~~\footnotesize{$\eta = 0.8$}&~~~~~~~~~~~~\footnotesize{$\eta = 1$} \\
\midrule
\multicolumn{8}{l}{\small{\textit{Example of cases where $\eta = 0.8$ succeeds and $\eta = 1$ fails. The network relies on relative location.}}}\\
\multicolumn{4}{c}{\includegraphics[width=0.48\linewidth]{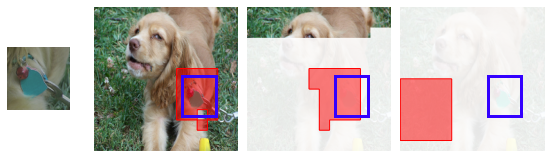}}
&
\multicolumn{4}{c}{\includegraphics[width=0.48\linewidth]{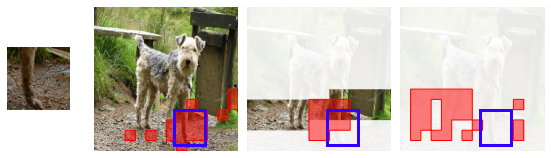}}
\\
\multicolumn{4}{c}{\includegraphics[width=0.48\linewidth]{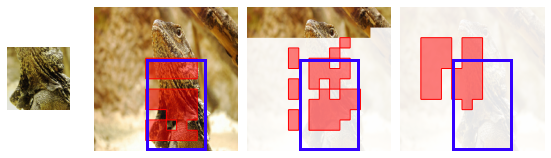}}
&
\multicolumn{4}{c}{\includegraphics[width=0.48\linewidth]{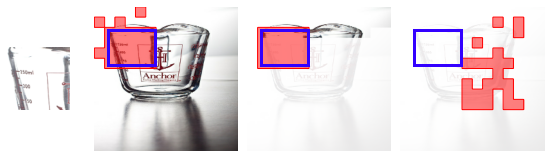}}
\\
\multicolumn{4}{c}{\includegraphics[width=0.48\linewidth]{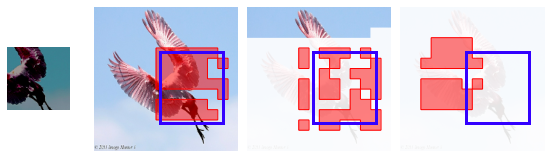}}
&
\multicolumn{4}{c}{\includegraphics[width=0.48\linewidth]{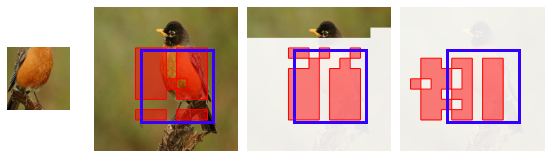}}
\\
\midrule
\multicolumn{8}{l}{\small{\textit{Example of cases where the query location can easily be inferred from looking at query alone.}}}\\
\multicolumn{4}{c}{\includegraphics[width=0.48\linewidth]{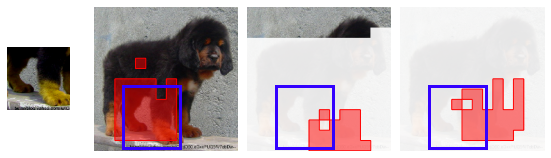}}
&
\multicolumn{4}{c}{\includegraphics[width=0.48\linewidth]{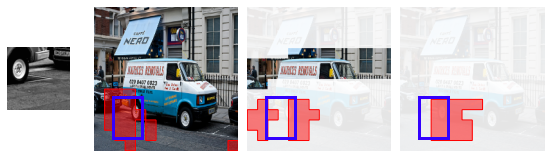}}
\\
\multicolumn{4}{c}{\includegraphics[width=0.48\linewidth]{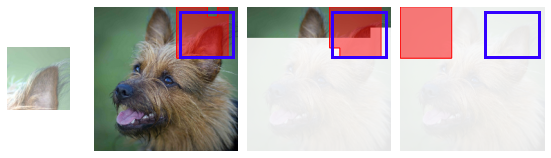}}
&
\multicolumn{4}{c}{\includegraphics[width=0.48\linewidth]{figures/images/download-16.png}}
\\
\midrule
\multicolumn{8}{l}{\small{\textit{Example of cases where the network relies on low-level cues. Only the variant with $\eta = 0$ succeeds.}}}\\
\multicolumn{4}{c}{\includegraphics[width=0.48\linewidth]{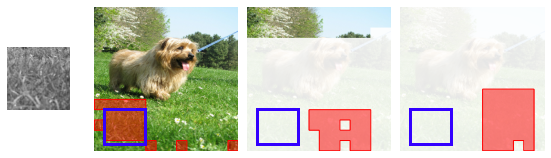}}
&
\multicolumn{4}{c}{\includegraphics[width=0.48\linewidth]{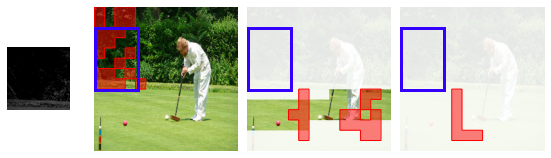}}
\\
\multicolumn{4}{c}{\includegraphics[width=0.48\linewidth]{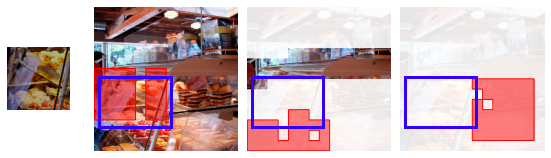}}
&
\multicolumn{4}{c}{\includegraphics[width=0.48\linewidth]{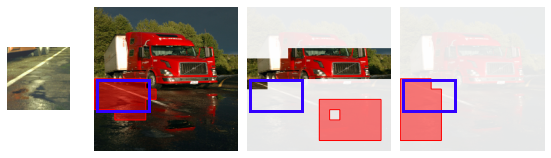}}
\\
\end{tabular}
\caption{
  \textbf{Visualizing LOCA's position predictions.}
  The query location is shown in blue in the reference and \OURS predictions are shown in red.
  Columns correspond to different reference masking rates and we show only patches visible to the query when it makes its prediction.
  Displayed images are not seen during training. See discussion in Section~\ref{ap:vis}.
}
\label{fig:visu_app}
\end{figure*}

In this section, we visualize the output of the position prediction pretraining task.
Specifically, in Figure~\ref{fig:visu_app}, we visualize query location prediction for different \OURS models.
We compare models pretrained with different masking rates:
(i)~$\eta=0$:~no masking, the reference is entirely visible to the query;
(ii)~$\eta=0.8$:~default masking rate, only 40 reference patch tokens are visible to the query;
(iii)~$\eta=1$:~full masking, the reference is invisible to the query.

In the first rows of Figure~\ref{fig:visu_app}, we show examples where the network seems to effectively solve the task by \textit{relative location}.
In those cases, we observe that \OURS trained with masking rate $\eta = 0.8$ manages to locate the query based on the patches visible from the reference.
For example we see that the network successfully manages to locate the leash joint based on seeing the patch representations of the head of the dog, or to locate the neck of the lizard based on the visible patches of its head.
By contrast, the network which does not see the reference at all (i.e. $\eta = 1$) cannot successfully locate the query in those cases.
Interestingly, we see that in some cases, this network ($\eta = 1$) can still locate the query by learning where things are typically located in natural images.
For example, we observe in Figure~\ref{fig:visu_app} that by recognizing a part such as ``ear'' it makes a guess that it is more likely to be at the top of the image rather than at the bottom.
However, it cannot guess if it is left or right because we apply random horizontal flips between query and reference during training and so this patch is as likely to occur at the right than at the left of the image.

Lastly, we observe that the network trained with full access to the reference ($\eta = 0$) can almost always locate the query.
This is because it can rely on low-level cues such as edge consistency or salient points.
The last rows of Figure~\ref{fig:visu_app} illustrate this phenomenon.

\end{document}